\documentclass[letterpaper, 10 pt, conference, twoside]{ieeeconf}
\usepackage{times}
\usepackage[pdftex]{graphicx}
\usepackage{subfig}
\usepackage{amsmath,amssymb,amsopn,amstext,amsfonts}
\usepackage{cancel}
\usepackage[space]{cite}
\usepackage{pdfsync}
\usepackage{balance}
\usepackage{color}
\usepackage{mathtools}
\usepackage{algorithm}
\usepackage{algorithmic}
\usepackage{bm}

\usepackage{diagbox}
\usepackage{booktabs}
\usepackage{paralist}
\usepackage[linkcolor=black,citecolor=black,urlcolor=black,colorlinks=true]{hyperref}

\makeatletter
\def\tagform@#1{\maketag@@@{\normalsize(#1)\@@italiccorr}}
\makeatother

\graphicspath{{./Figures/}}
\DeclareGraphicsExtensions{.pdf,.png,.jpg,.eps,.svg}
\IEEEoverridecommandlockouts
\title{DIABLO: A 6-DoF Wheeled Bipedal Robot Composed Entirely of Direct-Drive Joints}
\author{Dingchuan Liu$^{*1}$, Fangfang Yang$^{*1}$, Xuanhong Liao$^{*2}$, Ximin Lyu$^1$%
     \thanks{This work is supported by the Guangdong-Hongkong-Macao Joint Research of Science and Technology Planning Funding from Guangdong Province under Grant: 2023A0505010019.}
     \thanks{$^1$School of Intelligent Systems Engineering, Sun Yat-sen University, Guangzhou, China. $^2$Direct Drive Technology Limited, Dongguan, China. $^*$Indicates equal contribution.(\textit{Corresponding author:} Ximin Lyu)}
     \thanks{Email: {\tt\small lvxm6@mail.sysu.edu.cn}}
}
\begin{document}
\maketitle
\begin{abstract}
Wheeled bipedal robots offer the advantages of both wheeled and legged robots, combining the ability to traverse a wide range of terrains and environments with high efficiency. However, the conventional approach in existing wheeled bipedal robots involves motor-driven joints with high-ratio gearboxes. While this approach provides specific benefits, it also presents several challenges, including increased mechanical complexity, efficiency losses, noise, vibrations, and higher maintenance and lubrication requirements.

Addressing the aforementioned concerns, we developed a direct-drive wheeled bipedal robot called DIABLO, which eliminates the use of gearboxes entirely. Our robotic system is simplified as a second-order inverted pendulum, and we have designed an LQR-based balance controller to ensure stability. Additionally, we implemented comprehensive motion controller, including yaw, split-angle, height, and roll controllers. Through experiments in both simulations and real-world prototypes, we have demonstrated that our platform achieves satisfactory performance.
\end{abstract}

\section{INTRODUCTION}
Mobile robotics has gained increasing attention in recent years. Among the various configurations, legged robots have demonstrated reliability and robustness~\cite{SR_Anymal}. However, they are often limited in speed and face high transportation costs, particularly when carrying payloads. Conversely, purely wheeled robots do not have these drawbacks but may have limited off-road capabilities. While there are specialized configurations such as deformable wheels~\cite{Zheng_Lee_2019} and tracks designed to address these issues, they often compromise the robot's agility and speed.

Articulated wheel-legged robots offer a combination of advantages from both legged and wheeled robots. An example is ETH's quadruped ANYmal, which has a wheeled version demonstrating robust mobility~\cite{ETH-ANYmal-legged, ANYmal-wheeled-legged}. However, one drawback of such quadrupedal wheeled robots is the presence of numerous actuators, leading to increased costs and a larger footprint. To address this concern, wheeled bipedal robots (WBRs) have gained significant research attention. WBRs leverage the benefits of both wheeled and legged systems, mitigating the challenges associated with cost and footprint by utilizing fewer actuators. However, due to their underactuated nature, designing appropriate controllers is crucial to accomplish complex tasks in challenging terrains while maintaining balance.

In this work, we present DIABLO, a 6-DoF WBR composed entirely of direct drive joints. We accurately model its dynamics using a 2D simplified model and develop essential motion controllers, including a balance controller designed using LQR. Through simulation and real-world experiments, we validate the stability of our mechanical structure and control system.
\begin{figure}[t]
    \centering
    \includegraphics[width = 1.0\columnwidth]{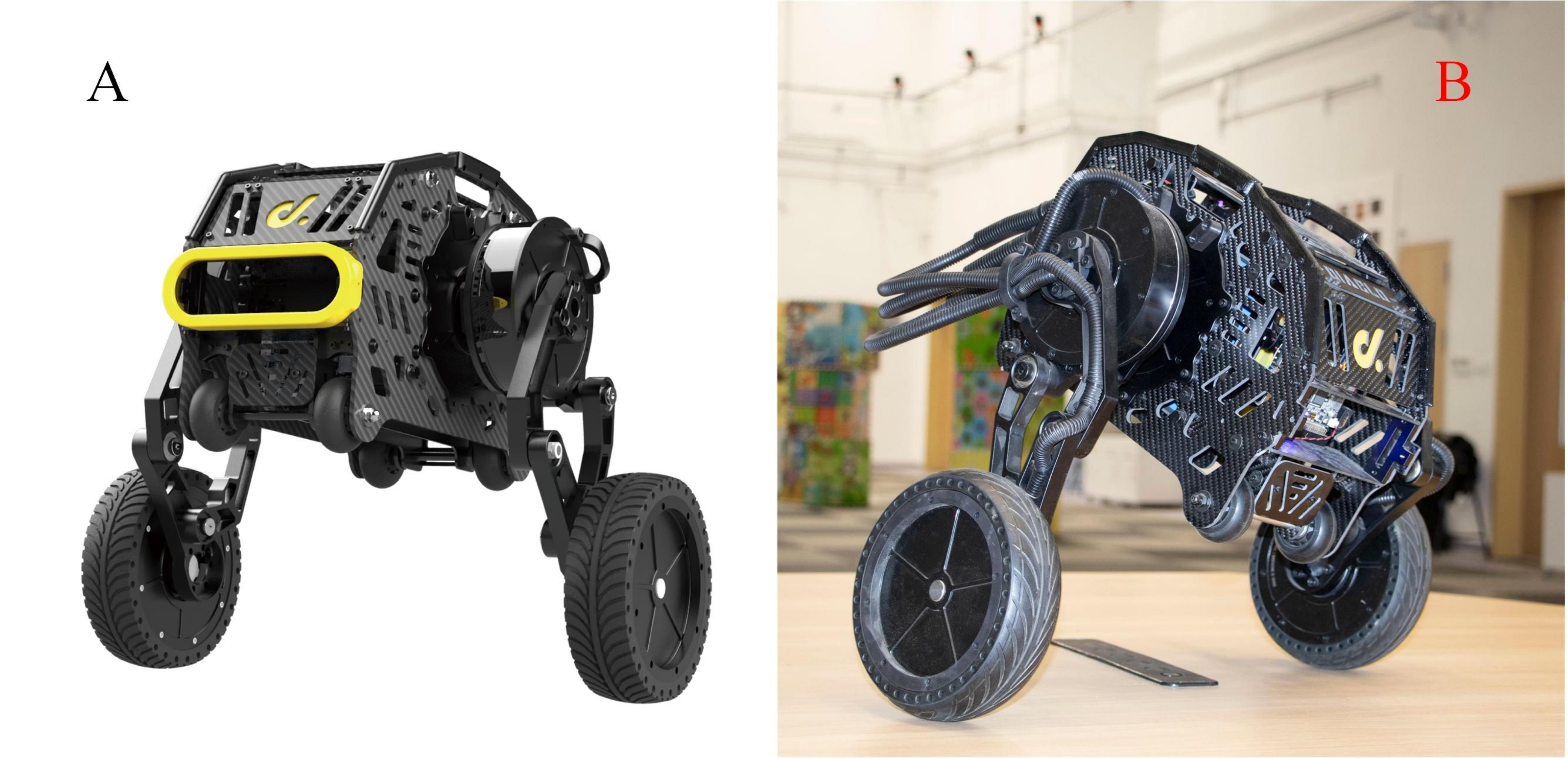}
    \caption{\textbf{A novel 6-DoF wheeled bipedal robot composed entirely of direct-drive joints.} (\textbf{A}) DIABLO keeping balance on flat ground. (\textbf{B}) DIABLO maintaining balance involving body tilt and non-horizontal orientation of the head. Accompanying video is available at https://youtu.be/mm0XruFvZ4U}
    \label{fig: A novel 6-DoF wheeled bipedal robot}
    \vspace{-0.7cm}
\end{figure}

There are three contributions of this paper:
\begin{enumerate}
    \item We present the mechanical, hardware, and software design of a novel WBR that is exclusively composed of direct-drive joints. This pioneering effort represents the first of its kind.
    \item We present a method applicable to various WBR: by simplifying the complex linkage structure of the legs and treating them as simple rods. Subsequently, we treat the model as a second-order inverted pendulum and perform individual dynamic analyses on the head, rod, and wheels. Additionally, we employ LQR to devise a balance controller for this simplified model, exhibiting excellent balance performance within approximately 30 degrees of the equilibrium point.
    \item We develop a comprehensive motion control method that enables the WBR to maintain balance even in challenging scenarios involving body tilt and non-horizontal orientation of the head (e.g., roll and pitch are not near zero as shown in Fig.~\ref{fig: A novel 6-DoF wheeled bipedal robot}B). The robustness of our system has been extensively validated through experiments conducted in both simulations and real prototypes.
\end{enumerate}


\section{RELATED WORK}
\label{sec:related_work}
One of the most impressive aspects of the research is the different mechanical design styles, focused on the degrees of freedom(DoF) and leg configuration problems. The most widely known WBR is the Handle robot released by Boston Dynamics~\cite{Handle}, but unfortunately, there is no related paper to introduce it in detail. The only thing known about Handle is that it uses a tandem leg configuration with a hydraulic drive. ETH's Ascento robot features a leg configuration with four linkages~\cite{Ascento-2019, Ascento-2020-LQR-WBC}, with a 1-DoF leg configuration. However, a notable limitation of the 1-DoF leg is its inability to control the head pitch angle during movement, since its joint directly controls height rather than pitch angle, making the pitch angle nearly equal to the pendulum angle\cite{AscentoPro}. For example, during acceleration, the head must tilt downward, and when head-mounted sensors need to detect ground information, the fixed head limits the viewing angle. SR600~\cite{SR600-HIT-1,sr600-hit-2}, a hydraulically driven robot designed by Harbin Institute of Technology (HIT), utilizes a series leg configuration with 2 DoFs. Ollie~\cite{Ollie-2021-Nonlinear, Ollie-2021-RL, Ollie-2022-RL+WBC, Ollie-2023}, designed by Tencent RoboticX Lab, showcases impressive mobility with a parallel leg configuration which is 2-DoF as well, and it incorporates a tail joint beneath the body. The drawback of the 2-DoF leg is the inability to control the ankle joint for lateral translation movements. Hotwheel~\cite{HotWheel} and Nezha~\cite{Nezha_NKD}, incorporates an ankle joint in its leg structure, featuring a total of 3 DoFs. Transitioning to a 3-DoF leg configuration increases the number of actuators, potentially elevating loads on joint motors and current solutions, utilizing motors with gearboxes, will cause additional joint clearance and maintenance costs. It should be noted that the above discussion is limited to the leg configuration and does not include the wheels. If the wheels are included, an additional DoF needs to be considered. After evaluating different leg configurations, we decided to remove the ankle joint and select a series 2-DoF configuration for DIABLO's leg, as it allows to control the head angle while maintaining a moderate number of actuators.

Besides the DoF and leg configuration problems, there are many control strategies to consider. Currently, available research programs focus on the robot balancing question. The reinforcement learning (RL) method has gained popularity~\cite{Ollie-2022-RL+WBC, Learing-WBR-CCC} yet grapples with the twin challenges of neural network non-interpretability and prolonged parameter training times. As for model-based force control algorithms, a simplified model analysis~\cite{HotWheel, Nezha_NKD} or a whole-body dynamics analysis~\cite{Ascento-2020-LQR-WBC} is generally required. After obtaining the dynamics equations of the robot, the controller design is typically carried out using LQR~\cite{Ascento-2020-LQR-WBC} or MPC~\cite{MPC-simulation-only}. Given the interpretability, generalizability, and ease of parameter tuning of the model-based approach in intricate environments, we've opted for this method in designing the DIABLO controller.

Additionally, given the implementation of torque control on the joints, the choice of actuator becomes crucial. It is pivotal for addressing the matching issues between the model and reality, as well as determining the upper limit of the robot's motion capabilities. As many legged robots opt for motors with gearboxes to enhance torque capacity, they often encounter issues like significant gearbox damage\cite{Mini-Cheetah}\cite{Mini-cheetah-3}. Conversely, AGVs typically employ direct-drive motors due to their high control bandwidth and minimal losses. To mitigate motor maintenance expenses and enhance joint control performance, we choose direct-drive motors as joints for DIABLO.


\section{SYSTEM DESCRIPTION}
\label{sec:system description}
\subsection{Mechanism}
\label{subsec:Mechanism}
\begin{table}[t]
    \centering
    \normalsize
    \caption{Specification of DIABLO}
    \begin{tabular}{|c|c|c|c|}
    \hline
        Total robot mass & 22.90~kg & $L_1$ & 0.090~m \\\hline
        Width & 0.540~m & $L_2$ & 0.140~m\\\hline    
        Length & 0.371~m & $L_3$ & 0.140~m\\\hline    
        Height when standing & 0.491~m & $L_4$ & 0.1400~m\\\hline
        Height when creeping & 0.260~m & $R$ & 0.0935~m\\\hline
    \end{tabular}
    \label{tab:Parameters of DIABLO mechanics}
    \vspace{-0.0cm}
\end{table}
\begin{figure}[t]
    \vspace{-0.0cm}
    \centering
    \includegraphics[width=1.0\columnwidth]{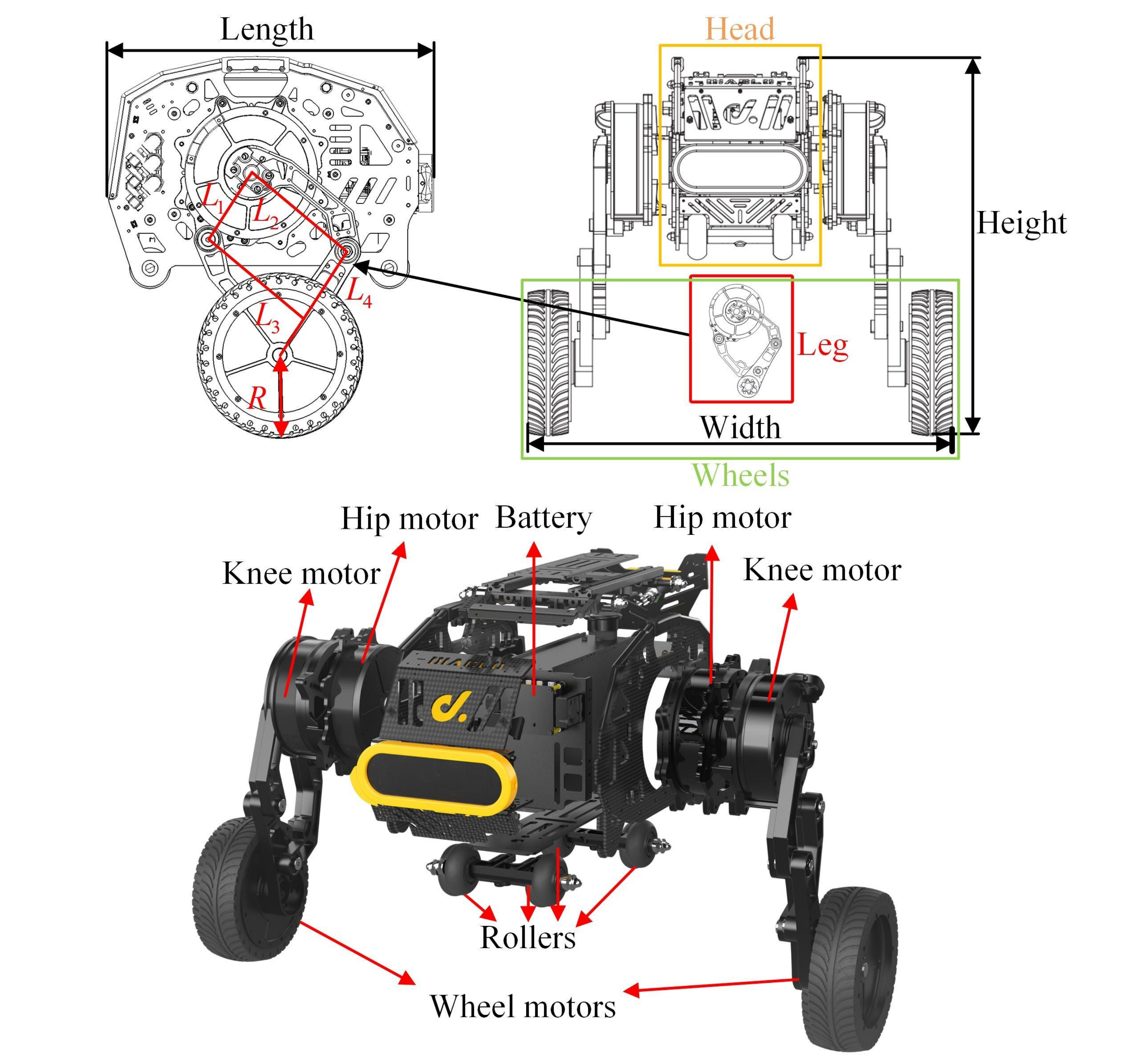}
    \caption{Mechanical structure of DIABLO}
    \label{fig:Mechanical structure of DIABLO}
    \vspace{-1.2cm}
\end{figure}
DIABLO features a total of 6 DoFs, with 2 DoFs dedicated to the wheels, while the remaining 4 DoFs are allocated to the hip and knee joints. All joints are efficiently powered by M1502D direct-drive motors. The implementation of direct-drive technology not only ensures a longer lifespan but also enhances system stability and improves overall reliability by eliminating transmission structure losses. The adoption of flange connections between linkages and joints simplifies the maintenance and repair processes of the robot, making it easily accessible for necessary adjustments. DIABLO can switch between crawling and standing, and the rollers under the head are used to assist the head movement during crawling.
Furthermore, the direct-drive solution eliminates friction in transmission mechanisms, resulting in higher power efficiency. 
With enhanced endurance, the robot can dedicate more time to tasks, reducing the frequency of recharging. More detailed information about the M1502D motor will be discussed in Subsection~\ref{subsec:Hardware}.

DIABLO's mechanical structure comprises three main components: the head, legs, and wheels. The head houses crucial components such as the battery, main control board, IMU, and more. Positioned beneath the head are four rollers (as depicted in Fig.~\ref{fig:Mechanical structure of DIABLO}), enabling DIABLO to switch to a crawling mode for forward movement. In this mode, the four rollers support the entire system, while the two wheels provide power, resulting in enhanced energy efficiency on horizontal terrains.


The hip motor is flanged to the knee motor, resulting in a series connection that requires less torque from the motors when the same force is applied at the foot, as compared to a parallel coaxial arrangement. The knee motor directly drives the leg components. DIABLO's leg configuration utilizes a parallelogram linkage, which plays a crucial role in distributing the forces acting on the linkages. Detailed parameters for each linkage can be found in Fig.\ref{fig:Mechanical structure of DIABLO} and Tab.\ref{tab:Parameters of DIABLO mechanics}. It is worth noting that some linkages are designed in an arched form to accommodate practical wiring considerations in the actual robot. During the analysis, these arched linkages can be treated as straight bars. Beneath the leg, the wheel motor is flanged to the 4th rod ($L_4$) of the leg.

The 2-DoF leg offers the capability to adjust both the height and horizontal angle of the robot's head. This feature allows for fine-tuning the overall center of mass (CoM), ensuring balance and stability even when carrying heavy cargo. In our tests, DIABLO demonstrated the ability to carry a weight of 4 kg in a standing position and up to 80 kg in crawling mode. It can operate continuously for up to 3 hours at an ambient temperature of 34 $^\circ$C. On flat ground, it can easily traverse obstacles of up to 8 cm in height. The head can be adjusted to a maximum height of 20 cm, enabling the robot to cross over a wide range of obstacles. The noise level during operation is below 49 dB, ensuring a relatively quiet performance.


\subsection{Hardware}
\label{subsec:Hardware}
\begin{figure}[t]
	\vspace{0.0cm}
	\centering
	\includegraphics[width=1.0\columnwidth]{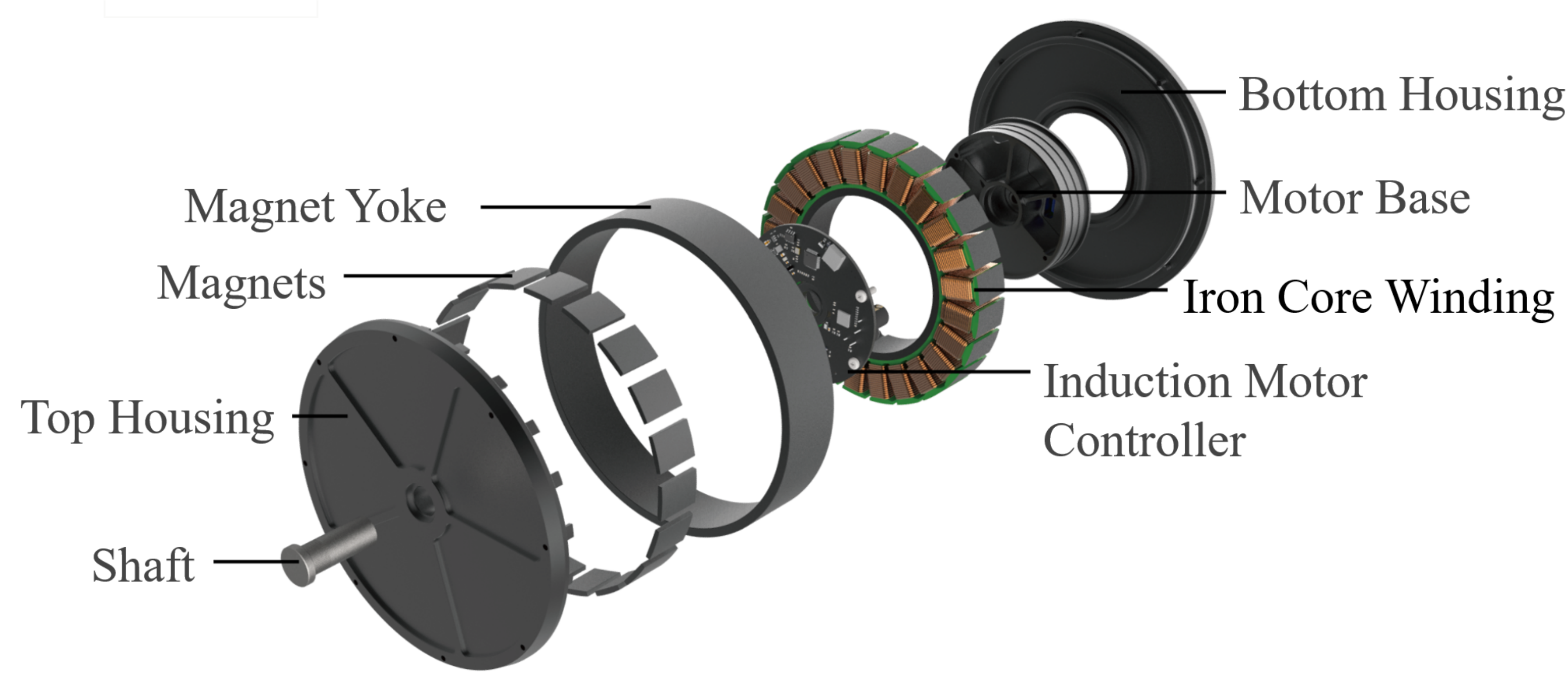}
	\caption{The exploded parts diagram of M1502D motor
	\label{fig: The exploded parts diagram of M1502D}}
	\vspace{-1.1cm}
\end{figure}
\begin{table}[t]
    \centering
    \normalsize
    \caption{Parameters of M1502D motor}
    \begin{tabular}{|c|c|}
    \hline
    Noload Bus Current & $\leq$ 0.18~A \\\hline 
    Rated Bus Current & 12~A \\\hline
    Maximum Efficiency & $\geq$ 76~$\%$ \\\hline
    Stall Current & 15.5~A \\\hline
    Operating Ambient Temperature & -15~$^\circ$C $\sim$ 45~$^\circ$C \\\hline
    Motor Weight & 2.6 $\pm$ 0.1~kg \\\hline
    Encoder Resolution & 16384\\\hline
    Absolute Accuracy & 8196\\\hline
    No-load Speed & 220~rpm \\\hline
    Rated Speed & 115~rpm\\\hline
    Rated Torque & 9.6~Nm \\\hline
    Rated Voltage & 24~V\\\hline
    Stall Torque & 17~Nm\\\hline
    \end{tabular}
    \label{tab:Parameters of M1502D motor}
    \vspace{-0.7cm}
\end{table}

To achieve exceptional force control performance, all joints of the DIABLO robot are powered by our advanced direct-drive motor, the M1502D. This motor has undergone meticulous optimizations in terms of pole number, slot type, and the utilization of specially formulated permanent magnet materials. These optimizations ensure the motor delivers a wide range of rotation speeds and high torque output while effectively minimizing torque fluctuations.
The motor drive system employs the Field-Oriented Control (FOC) algorithm along with a high-precision angle sensor integrated into the motor itself. This combination enables precise control of motor torque, position, and speed, ensuring excellent quietness in operation. Additionally, the drive system incorporates comprehensive motor On-Board Diagnostics (OBD) monitoring mechanisms and protective functions, ensuring the motor operates safely and reliably.
For a detailed illustration of the exploded parts of the motor, please refer to Fig.~\ref{fig: The exploded parts diagram of M1502D}, and for specific parameters, please consult Tab.~\ref{tab:Parameters of M1502D motor}.

DIABLO's motion control system is effectively managed by two main boards: a feature-rich micro-controller and a high-performance mini PC. The micro-controller board utilizes the CAN bus to establish communication with the motor drive board. It offers a wide range of I/O interfaces for serial communication, enabling functionalities such as remote control and firmware updates. Additionally, the micro-controller board integrates the BMI088, a high-performance IMU that combines a triaxial accelerometer and a triaxial gyroscope. This IMU communicates with the micro-controller board through the SPI interface, facilitating precise motion sensing and control.

On the other hand, the mini PC is equipped with a powerful 4-core ARM A53 processor, providing 5 TOPS edge inference capabilities. It supports multiple camera sensor inputs and features H.264/H.265 encoding/decoding capabilities. Communication between the micro-controller and the mini PC is established through the UART interface, enabling seamless coordination and data exchange between the two boards.

\subsection{Software}
\label{subsec:Software}
\begin{figure}[t]
    \centering
    \includegraphics[width = 1.0\columnwidth]{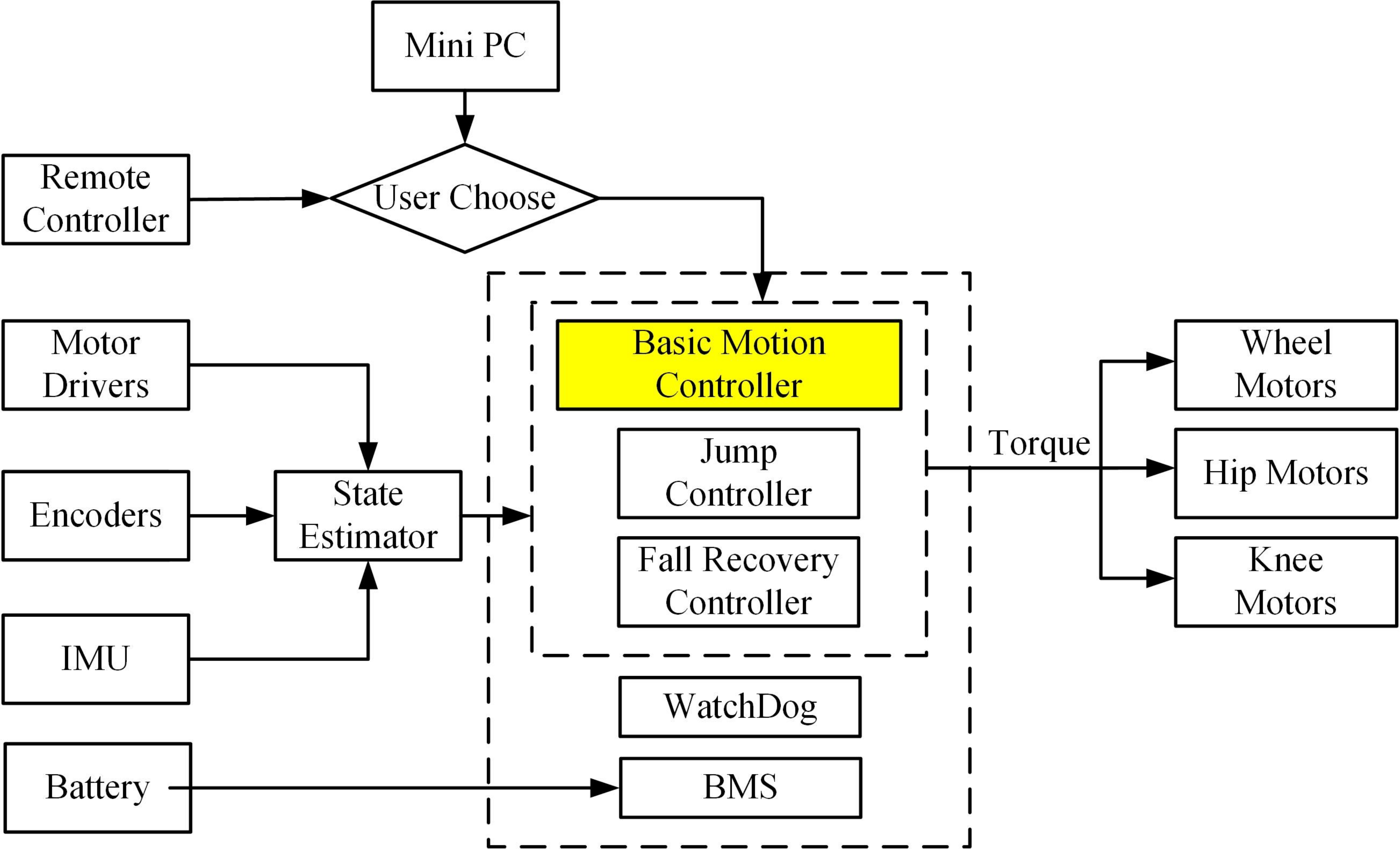}
    \caption{Overview of the software framework}
    \label{fig:overview of the software framework}
    \vspace{-0.0cm}
\end{figure}

The micro-controller in DIABLO serves a dual purpose: handling peripheral communication and playing a crucial role for the motion control of DIABLO. It incorporates the ChibiOS/RT operating system and runs the basic controller, jump controller, and fall recovery controller. The basic controller includes balance controller and comprehensive motion controller, which are introduced in Sec.~\ref{sec:controller design}.
Additionally, we have developed an open-source ROS framework~\cite{DIABLO_ROS} on the mini PC. This framework enables users to control the robot for navigation, recognition, and other computationally intensive tasks. Its effectiveness in enhancing the robot's capabilities has been demonstrated and referenced in~\cite{Fast-DIABLO}.


The software framework is illustrated in Fig.~\ref{fig:overview of the software framework}. A complementary filter based state estimator is established to estimate the required states by utilizing information acquired from the IMU, motor encoder, and motor drivers.
The controller reference signal can originate from either the mini PC or the remote controller. Both the reference signal and current states are sent to the basic motion controller, jump controller, and fall recovery controller, respectively. The designed torque is calculated and sent to the motor drivers of the wheel, hip, and knee motors.
DIABLO also incorporates a battery management system (BMS) to optimize battery utilization. Furthermore, a Watchdog system is implemented to detect software anomalies and ensure system reliability and safety.
In this paper, we will provide detailed information about the basic motion controller. The jump controller and fall recovery controller are demonstrated in the attached video.
 
\section{MODELING}
\label{sec:modeling}
\begin{figure}[t]
	\vspace{0.0cm}
	\centering
	\includegraphics[width=1.0\columnwidth]{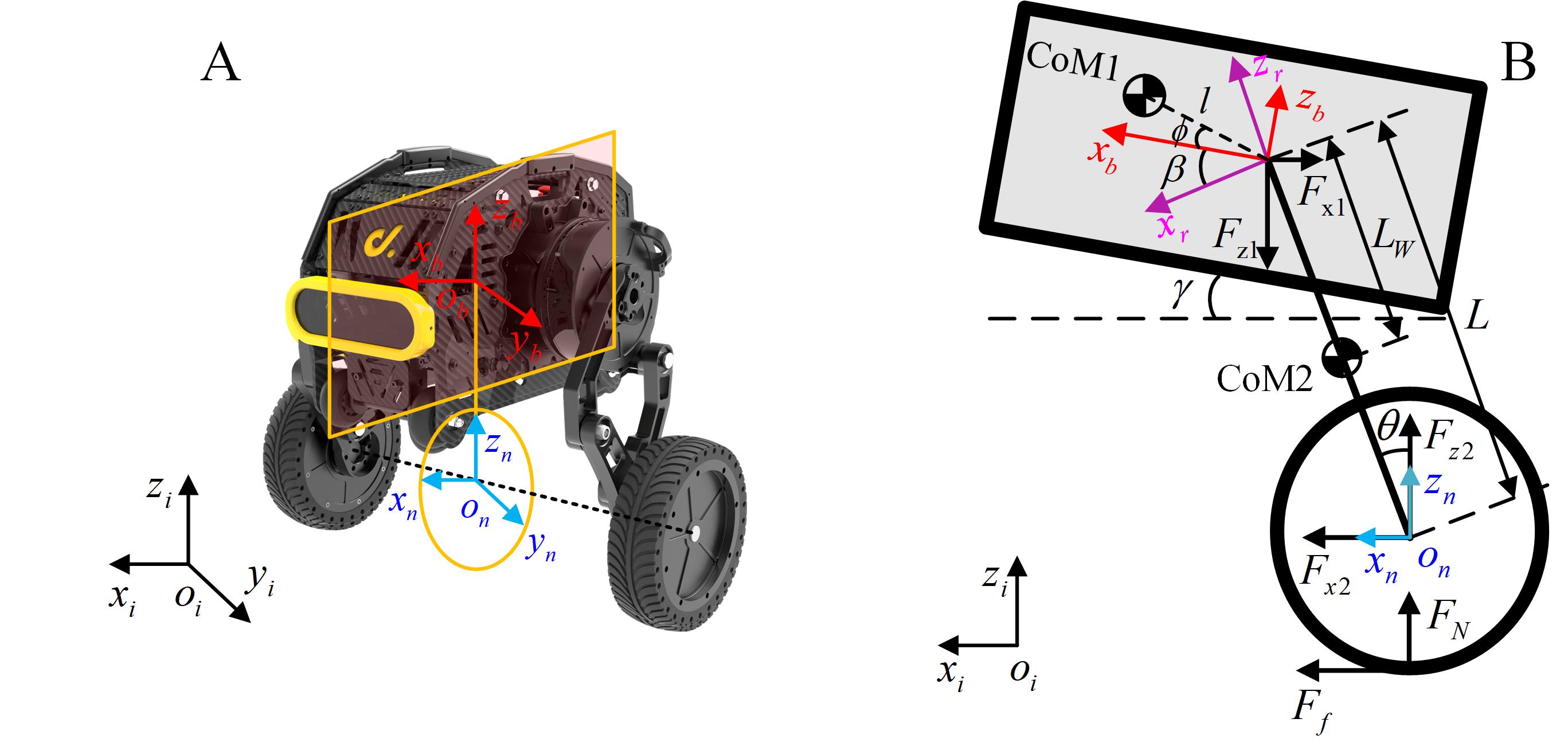}
	\caption{{\textbf{DIABLO's coordinate system and dynamic analysis diagram of simplified model} (\textbf{A}}) DIABLO's coordinate system. (\textbf{B}) Illustration of the simplified model in a 2D planar environment along with the dynamic analysis diagram.}
	\label{fig: Dynamic analysis diagram of simplified model}
 \vspace{-0.8cm}
\end{figure}

As illustrated in Fig.~\ref{fig: Dynamic analysis diagram of simplified model}, a local control frame ($o_n$) is positioned at the midpoint of the line connecting the two wheels. The body frame ($o_b$) and simple rod frame ($o_r$) are established relative to the local control frame. The body frame's origin is positioned at the center of the IMU, which is installed inside the head. Specifically, the IMU is situated at the midpoint of the centerline connecting the left and right hip joints. Prior to employing the Newton-Euler method for analysis, the leg linkage structure was initially simplified to a variable-length straight rod, and the model was further reduced to a 2D plane. The pitch, yaw, and roll values are obtained from the rotation of the body frame in relation to the inertia frame.

As shown in Fig.~\ref{fig: Dynamic analysis diagram of simplified model}B, the CoM1 represents the center of mass of the simplified head. The position of CoM1 can be expressed as:
\begin{equation}\label{eq: CoM_of_head}
\begin{aligned}
    {}_Ix_H &= {}_Ix + Lsin\theta + lcos(\beta + \theta + \phi)\\
    {}_Iz_H &= Lcos\theta + lsin(\beta + \theta + \phi)
\end{aligned}
\end{equation}
where ${}_Ix$ is the displacement of the $o_n$ system relative to the $o_i$ system in the x-direction; $L$ is the length of the rod; $\theta$ is the pendulum angle of the rod; $l$ is the distance from the CoM1 to the hip joint center; $\beta$ represents the angle of rotation of the $o_b$ frame relative to the $o_r$ frame; $\phi$ is the angle between the CoM1 and the head's x-axis of rotation($x_b$).

The CoM2 represents the center of mass of the simplified rod. The position of CoM2 can be expressed as:
\begin{equation}\label{eq: CoM_of_rod}
    \begin{aligned}
        {}_Ix_L &= {}_Ix + L_Msin\theta\\
        {}_Iz_L &= L_Mcos\theta
    \end{aligned}
\end{equation}
where $L_M$ is the distance from the CoM2 to the wheel center, and $L_M$ satisfies the equation $L_M = L - L_w$. $L_w$ is the distance from the hip joint center to CoM2.

The wheel is analyzed to obtain as following:
\begin{equation}\label{eq: wheels}
    \begin{aligned}
    F_f &= m_w\Ddot{{}_Ix}+F_{x2}\\
    F_N &= m_wg+F_{z2}\\
    I_w\frac{\Ddot{{}_Ix}}{R} &= \tau_w-F_fR  
    \end{aligned}
\end{equation}
where $F_{f}$ and $F_N$ denote the horizontal and vertical component forces, respectively, at the point of contact between the ground and the wheel; $F_{x2}$ and $F_{z2}$ are the horizontal and vertical components of the force on the rod at the connection with the wheel respectively; $m_w$ is the mass of the wheel; $I_w$ is the inertia about the CoM of the wheel; $\tau_w$ is the torque of the wheel motor; $R$ is the radius of the wheel; $g$ is the gravity.

The rod is analyzed to obtain as following:
\begin{equation*}\label{eq: rod_x}
      F_{x2} = m_{L}\frac{d^2}{dt^2}({}_Ix_L) + F_{x1}
\end{equation*}
\begin{equation}\label{eq: rod_y}
    F_{z2} = m_L\frac{d^2}{dt^2}({}_Iz_L)+ F_{z1}+ m_{L}g
\end{equation}
\begin{align*}\label{eq: rod_omega}
   I_L\ddot{\theta} = -\tau_w - \tau_H &+(F_{z1}\sin\theta - F_{x1}\cos\theta)L_w\\
    &+ (L-L_w)(F_{z2}\sin\theta - F_{x2}\cos\theta)
\end{align*}
where $Fx1$ and $Fz1$ are the horizontal and vertical components
of the force on the top end of the rod respectively; $m_L$ is the mass of the total mass of leg linkages; $\tau_H$ is the torque of the hip motor.

The head is analyzed to obtain as following:
\begin{equation*}\label{eq: head_x}
    F_{x1} = m_H\frac{d^2}{dt^2}({}_Ix_H)
\end{equation*}
\begin{equation}\label{eq: head_y}
    F_{z1} = m_H\frac{d^2}{dt^2}({}_Iz_H) + m_Hg
\end{equation}
\begin{equation*}\label{eq: head_omega}
    I_H(\ddot{\theta}+\ddot{\beta}) = \tau_H -F_{z1}lcos(\theta+\beta+\phi)-F_{x1}lsin(\theta+\beta+\phi)
\end{equation*}
where $m_H$ is the mass of the head. $I_w$, $I_L$, and $I_H$  are all obtained from the SolidWorks model except $I_L$, which can be derived from the parallel axis theorem $I_d = I + md^2$.

After solving the equations of \eqref{eq: wheels}-\eqref{eq: head_y} and linearizing them around upward equilibrium, we can get the equation of motion(EoM) for the 2D simplified model:
\begin{align}\label{eq: equation_of_motion}
    \tau_H &= I_H(\ddot{\beta} + \ddot{\theta})  \nonumber\\
    \tau_w &= I_H\ddot{\beta} + (I_H+I_L+L^2m_H+ L^2m_L + L_w^2m_L \nonumber\\
    &- 2LL_wm_L)\ddot{\theta}+(Lm_H+Lm_L-L_wm_L)\ddot{x} \nonumber  \\
    &(L_wm_Lg-Lm_Lg-Lm_Hg)\theta\\
    F_w &= (Lm_H+Lm_L-L_wm_L)\ddot{\theta}+(m_H+m_L+\frac{I_w}{r^2})\ddot{x}\nonumber 
\end{align}
The EoM of the DIABLO in the 2D plane can be written as:
\begin{equation}\label{eq: expression_for_the_dynamics}
    \bm{a\ddot{q} + bq = cu}\\
\end{equation}
where $\bm{q} = [\gamma, \theta, x]^T, \bm{u}=[\tau_H, \tau_w]^T$, $\gamma$ is the pitch angle of the head of DIABLO in the 2D model, which satisfies equation $\gamma = \beta + \theta$. The controlled state variable $\bm{q}$ must be obtained through the transformation of joint position state variables. The corresponding relationship is given by:
\begin{equation}\label{eq: trans_joint_2_controllerstate}
    \begin{bmatrix}
        \gamma\\\theta\\x
    \end{bmatrix} = \begin{bmatrix}
        1&0&0&0\\
        1&1&\frac{1}{2}&0\\
        R&R&R&R
    \end{bmatrix}\begin{bmatrix}
        \gamma\\\sigma_1\\\sigma_2\\\sigma_3
    \end{bmatrix}
\end{equation}
where $\sigma_{i} (i=1,2,3)$ stands for the angle of the hip motor, knee motor, and wheel motor, which are got by the encoders. The coefficient matrix of the equation of dynamics is deduced as follows:
\begin{equation}\label{eq: coeff_of_eq_for_dynamics}
    \bm{a} = \frac{\partial \bm{f}(EoM)}{\partial{\ddot{\bm{q}}}}, 
    \bm{b} = \frac{\partial \bm{f}(EoM)}{\partial\bm{q}}, 
    \bm{c} = \begin{bmatrix}
        -1&0\\
        0&-1\\
        0&\frac{1}{r}
    \end{bmatrix}
\end{equation}
Transform the ~\eqref{eq: expression_for_the_dynamics} to the state-space expression:
\begin{equation}
\bm{\dot{X} = AX+Bu}
\end{equation}
where 
\begin{equation}\label{eq: state_space}
    \begin{aligned}
    \bm{X} &= \begin{bmatrix}
    \gamma&\theta&{}_Ix&\dot{\gamma}&\dot{\theta}&\dot{{}_Ix}&\int\gamma dt & \int {}_Ix dt
    \end{bmatrix}^T \\
    \bm{u} &= \begin{bmatrix}
        \tau_H&\tau_w
    \end{bmatrix}^T\\
    \bm{A} &= \begin{bmatrix}
    \bm{0}_{3\times3}&\bm{I}_{3\times3}&\bm{0}_{3\times2}\\ 
    \bm{a}^{-1}b&\multicolumn{2}{c}{\bm{0}_{3\times5}}\\
    \begin{bmatrix}
        1&0&0\\
        0&0&1
    \end{bmatrix}&\multicolumn{2}{c}{\bm0_{3\times5}}
    \end{bmatrix},
    \bm{B} = \begin{bmatrix}
        \bm{0}_{3\times2}\\
        \bm{a^{-1}c}\\
        \bm{0}_{2\times2}
    \end{bmatrix} 
    \end{aligned}
\end{equation}
All states can be directly measured or analytically inferred through sensor readings, which means the output matrix of the system can be written as:
\begin{equation} \label{eq: output_matrix_of_system}
    \bm {Y} = \bm{C}\bm{X}, \bm{C} = \bm{I}_{8\times8}
\end{equation}
The determination of the state space model involves substituting model parameters into the equations to obtain the state matrix $\bm{A}$ and control matrix 
$\bm{B}$. The controllability matrix is of full rank, indicating that the system is controllable. Furthermore, the output matrix $\bm{C}$ is an identity matrix $\bm{I}_{8\times8}$, establishing the system's observability.

\section{CONTROLLER DESIGN}
\label{sec:controller design}
\begin{figure*}[t]
    \vspace{0.0cm}
    \centering
    \includegraphics[width=2.0\columnwidth]{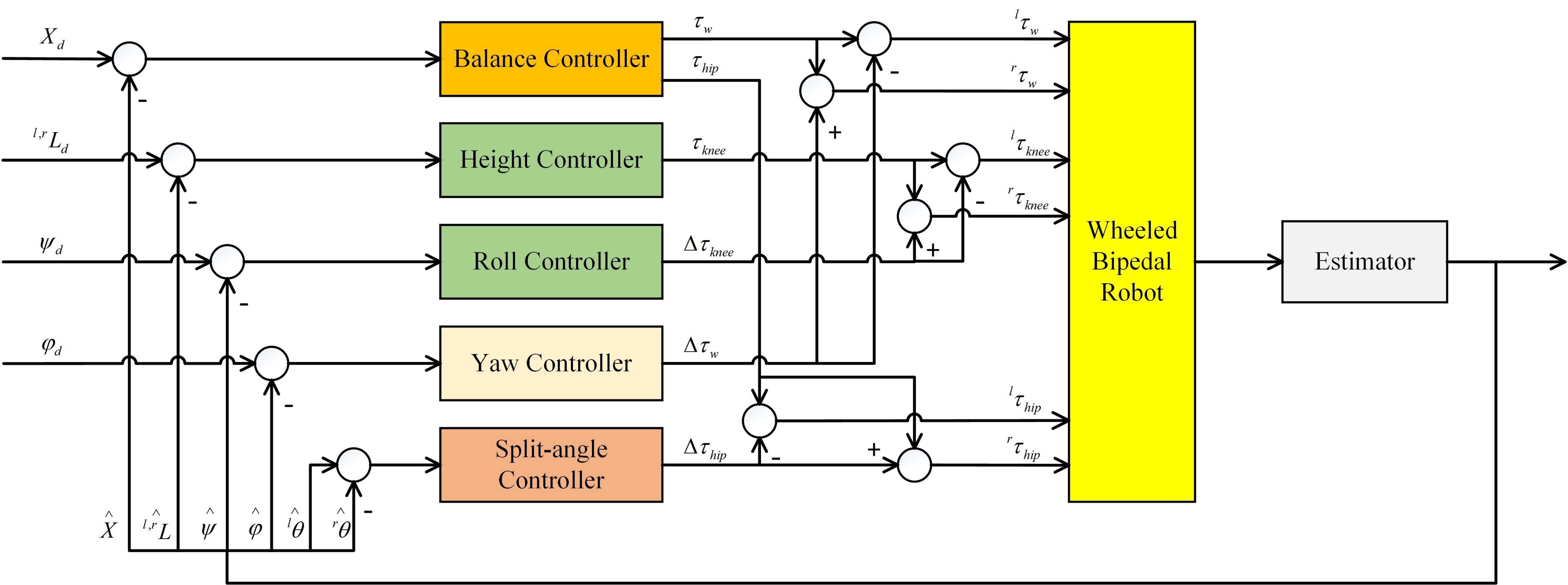}
    \caption{Basic Motion Controller of DIABLO
    \label{fig: Basic Motion Controller of DIABLO}}
    \vspace{-0.3cm}
\end{figure*}
As illustrated in Fig.~\ref{fig: Basic Motion Controller of DIABLO}, the basic motion control system of DIABLO comprises the balance controller and the comprehensive motion controller. Besides, the comprehensive motion controller consists of four key components: the height controller, roll controller, yaw controller, and split angle controller.

\subsection{Balance Controller Design}
Building upon the state-space model established in the previous section, for a multi-input multi-output system, we apply LQR to our model, which takes both regulation cost function and actuator output into performance and achieves some balance between them. By tuning the state penalty matrix $\bm{Q}$ and the control penalty matrix $\bm R$, we solve the Discrete-time Algebraic Riccati Equation(DARE) to determine a set of optimal feedback control gain matrices $\bm K$. Thus, we design the control law as a linear combination of the system states: 
\begin{equation}
    \bm{u = -K_{2\times8}X}
\end{equation}
The DARE is:
\begin{equation}
    \small\bm{P_{k+1} = A^T P_k A - (A^T P_k B) (R + B^T P_k B)^{-1} (B^T P_k A) + Q}
\end{equation}
As we have already obtain the continues-time state-space model, we can discretize it and find optimal solution by:
\begin{equation}
    \bm{K = (R+B^TPB)^{-1}B^TPA}
\end{equation}
By employing the aforementioned approach, system stability can be achieved in the vicinity of the linearized equilibrium point. To facilitate trajectory tracking for the robot, it is necessary to incorporate a reference input into the system's inputs, i.e.
\begin{equation}
    \begin{gathered}\label{eq: control_law}
    \bm{u = K(X_d - X)} \\
    \bm{X_d} = \begin{bmatrix}
        \gamma_d&0&{}_Ix_d&0&0&0&0&0
    \end{bmatrix}^{T}
    \end{gathered}
\end{equation}

Through this design, the balance controller based on LQR and 2D simplified model enables DIABLO to achieve excellent balance and straight motion.

\subsection{Comprehensive Motion Controller Design}
The comprehensive motion controller elaborates on the 2D simplified model, incorporating various fundamental motion modules, including height controller, roll controller, yaw controller, and split-angle controller. These modules constitute the foundational operational capabilities of a WBR.

Through the height controller, DIABLO can adjust its leg length to adapt to the controller's demands and different terrain. According to the geometric configuration of its legs, we can deduce the relationship between the leg length and the robot height with respect to the joint position state quantities:
\begin{equation}\label{eq: height_relationship_of_robot}
    L_{\text{leg}} = 2Lcos(\frac{\sigma_2}{2}), H = L_{\text{leg}}cos\theta
\end{equation}
The leg length satisfies the following kinetic relationship:
\begin{equation}\label{eq: tau_knee}
    \begin{gathered}
    \tau_{\text{knee}} = J^T(\Delta F + m_Hg) \\
    \Delta F = \ddot{L}_{\text{leg},d} + k_p(L_{\text{leg},d} - L_{\text{leg}}) + k_d(\dot{L}_{\text{leg}, d} - \dot{L}_{\text{leg}}) 
    \end{gathered}
\end{equation}
where $\tau_{\text{knee}}$ is the initial torque output acting on the knee motor; $\Delta F$ is the output force of the combined PD and feed-ward controller to adjust height changes in user input; $L_{\text{leg},d}$, $\dot{L}_{\text{leg}, d}$,$\Ddot{L}_{\text{leg},d}$ are the desired leg length, the desired rate of leg length change, and the desired acceleration of leg length change respectively; The $L_{\text{leg}}$ and the $\dot{L}_{\text{leg}}$ are the actual value of the leg length and the rate of its change respectively; the $k_p, k_d$ are the parameters of the PD controller. 

In the roll controller, the PD controller is utilized for the control of the body roll angle, which is given as 
\begin{equation} \label{eq: delta_tau_roll}
    \Delta \tau_{roll} = k_p(\psi_d-\hat{\psi})+k_d(\dot{\psi}_d-\hat{\dot{\psi}})
\end{equation}
where $\hat{\psi}$ and $\hat{\dot{\psi}}$ are derived from the estimator by the sensor data after fusion solving. The $\Delta \tau_{roll}$ is integrated with the $\tau_{\text{knee}}$ derived from the height controller to obtain the knee motors' torque input $^l\tau_{\text{knee}}$ and $^r\tau_{\text{knee}}$. By using this module, DIABLO can keep balance within a certain range of roll angles, and the head can even be non-horizontal. 

The yaw controller was devised utilizing LQR. The resulting output $\Delta \tau_{w}$ is integrated with the wheel output $\tau_w$ derived from the balance controller to get the wheel motors' torque input $^l\tau_w$ and $^r\tau_w$.

The split-angle controller is to address the model mismatch issue in our balance model. As our balance model relies on a 2D simplified model derived from a 3D model, the torque difference generated by the wheel motors during steering introduces a torque on the robot along the ground normal. This torque, responsible for steering the robot, also drives the robot's legs in the opposite direction, leading to a split. 
However, such a split angle still satisfies the balance condition when merged into the 2D simplified model. This mismatch arises due to the disparity between the theoretical balance control model and the actual situation. Applying PD controller to the angular difference $\delta\theta={}^{r}\theta -{}^{l}\theta$, we obtain the torque output required to maintain consistent angles. This torque is then superimposed with an opposite sign onto $\tau_H$ to derive the hip motors' torque inputs $^l\tau_{\text{hip}}$ and $^r\tau_{\text{hip}}$.

\section{EXPERIMENT VERIFICATION}
\label{sec:experiment_verification}
\subsection{Experiment in Simulation}

\begin{figure}[t]
    \centering
    \includegraphics[width = 1.0\columnwidth]{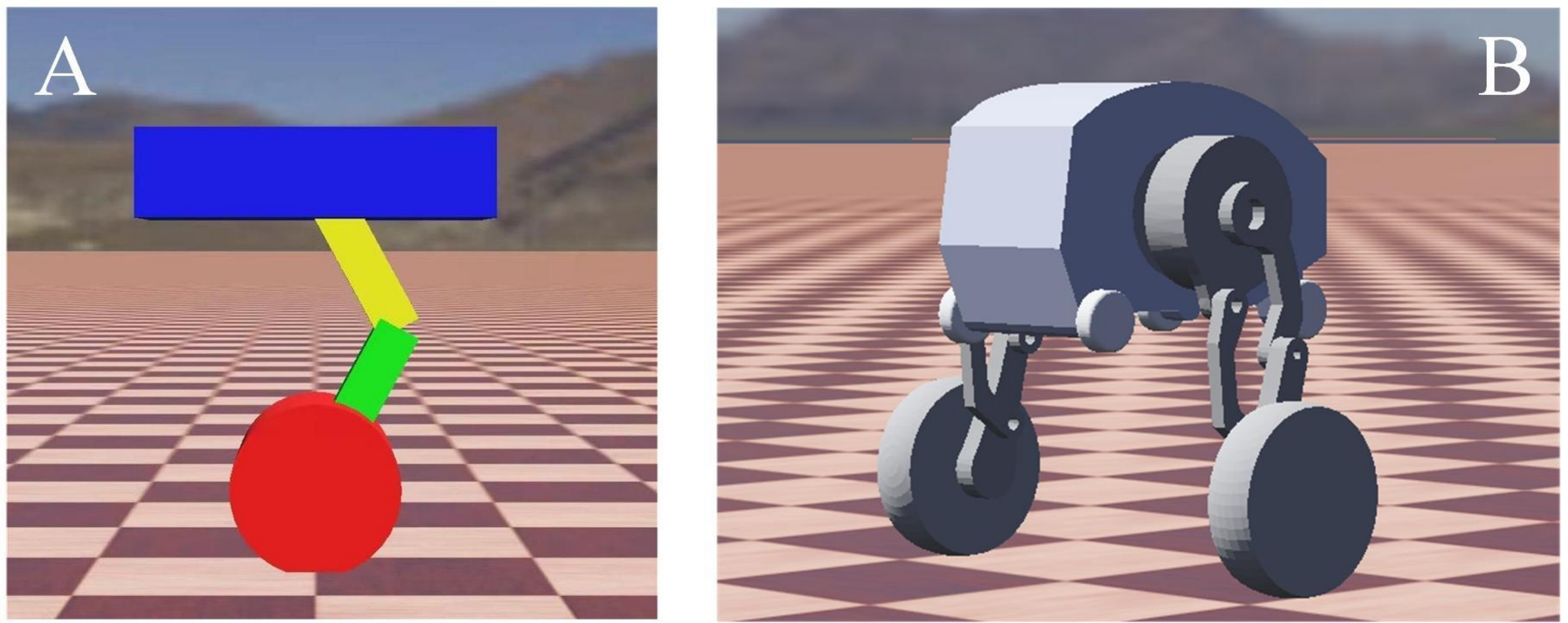}
    \caption{\textbf{DIABLO in simulation} (\textbf{A}) 2D simplified model (\textbf{B}) Complete DIABLO}
    \label{fig:DIABLO in simulation}


    \centering
    \includegraphics[width = 1.0\columnwidth]{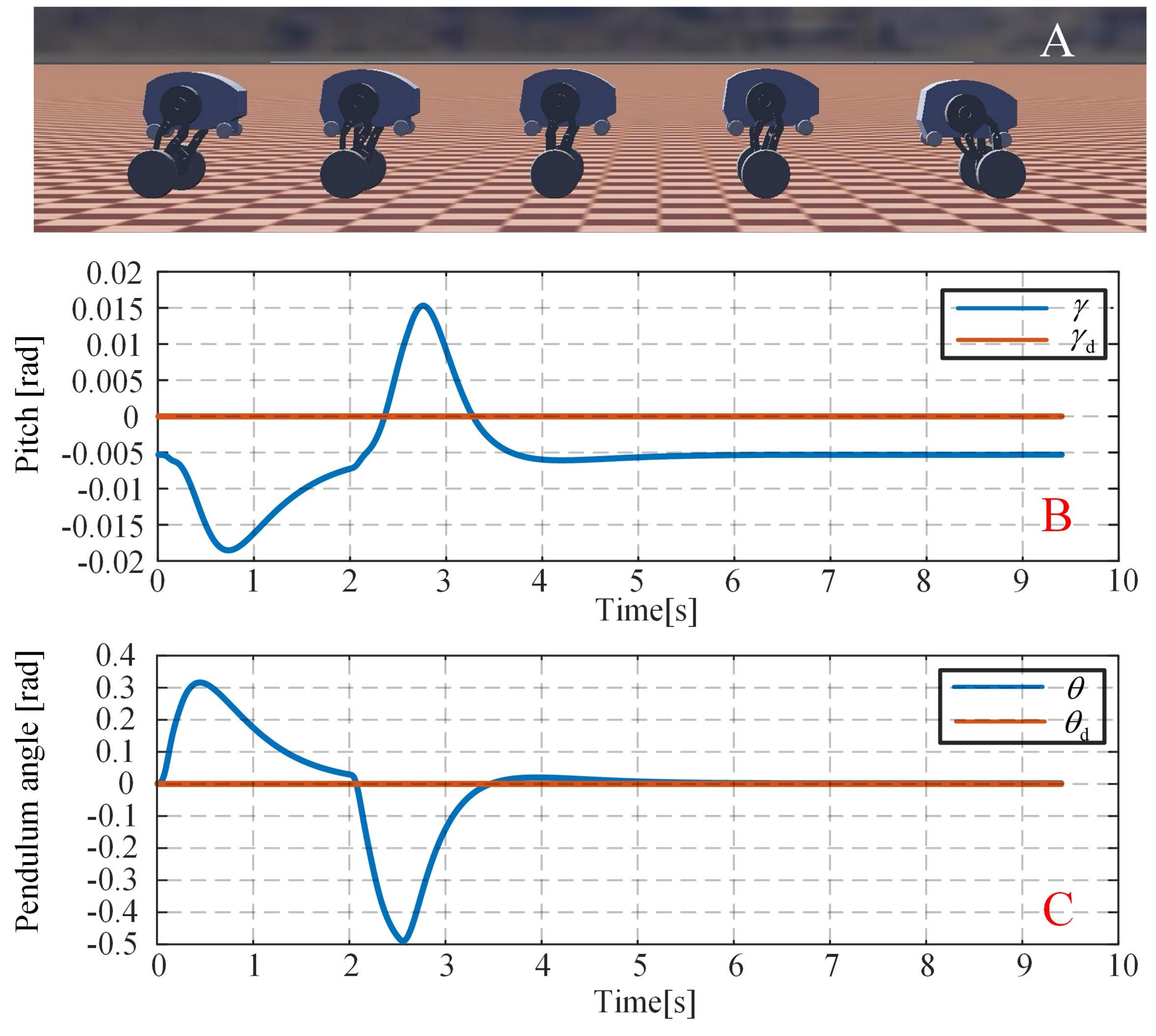}
    \caption{\textbf{Experiment of balance controller: straight motion on flat ground (in the simulation)} (\textbf{A}) DIABLO's pose performance during locomotion on flat ground in Webots, tracking a speed expectation of 2 m/s. (\textbf{B}) Pitch data during motion. (\textbf{C}) Leg's pendulum angle data during motion.}
    \label{fig:DIABLO'S attitude and data during motion in simulation}
    \vspace{-0.7cm}
\end{figure}

In order to verify the viability of our 2D simplified model controller, which is crucial for DIABLO's ability to maintain balance and motion, we employ Matlab as our code compiler and Webots as our simulation environment.
As depicted in Fig.~\ref{fig:DIABLO in simulation}A, we simplify the leg structure into a single rod for the simulation. This simplified representation has been thoroughly tested and validated to achieve a stable balance in the simulation.
LQR parameters for the balance controller and PID parameters for the height controller can be conveniently tuned in the simplified model simulation, providing a basis for subsequent simulations.

We then imported the complete model into Webots for simulation, as illustrated in Fig.~\ref{fig:DIABLO in simulation}B. Building upon the parameters fine-tuned in the previous stage, we sequentially add the split-angle controller, roll controller and yaw controller to complete the coarse tuning of all necessary control parameters. In the simulation, we conducted velocity and angular velocity tracking tests, where the robots effectively tracked the desired signals with high accuracy.

In Fig.~\ref{fig:DIABLO'S attitude and data during motion in simulation}, DIABLO tracks a velocity target of 2 m/s for a duration of 2 seconds. Throughout the motion, the leg's pendulum angle, deduced from (\ref{eq: trans_joint_2_controllerstate}), is consistently maintained within 0.5 rad, while the pitch angle of the head remains within 0.02 rad. These results unequivocally demonstrate the effectiveness of the controller design.
The successful simulation outcomes validate the robustness and reliability of our controller design, affirming its capability to achieve accurate tracking and maintain stability during dynamic motions in DIABLO.

\subsection{Experiment in Real-world Prototypes}
\begin{figure}[t]
    \centering
    \includegraphics[width = 1.0\columnwidth]{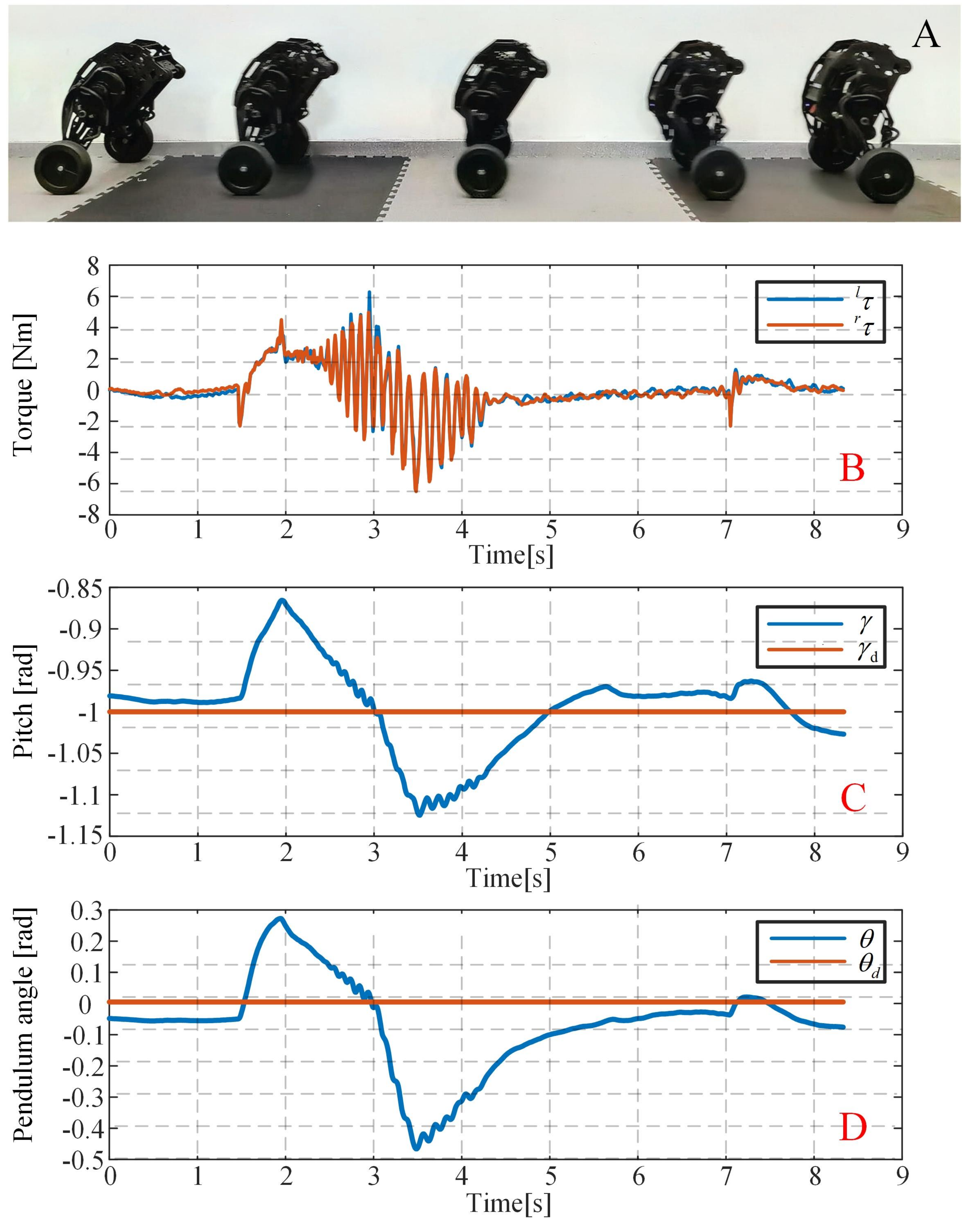}
    \caption{\textbf{Experiment of balance controller: straight motion on flat ground (in the real-world prototype)} (\textbf{A}) DIABLO's pose performance during locomotion on flat ground, tracking a speed expectation of 2 m/s. (\textbf{B}) Torque data of wheel motors during motion. (\textbf{C}) Pitch data during motion. (\textbf{D}) Leg's pendulum angle data during motion.}
    \label{fig:DIABLO'S attitude and data during motion in real-wold prototype}
    \vspace{-1.0cm}
\end{figure}

Upon deploying the controller parameters derived from previous simulations onto the real-world prototype, only minor adjustments were required to achieve optimal performance. In the experiment, we aimed to demonstrate the superior control of the pitch angle using our balance controller in a more intuitive manner. To achieve this, we maintained the pitch angle at 1.0 rad while simultaneously tracking a desired speed of 2 m/s. As depicted in Fig~\ref{fig:DIABLO'S attitude and data during motion in real-wold prototype}, the pitch angle during the motion remains consistently within the range of 1.0 ± 0.15 rad. Simultaneously, the leg pendulum angle is within the range of -0.5 rad to 0.3 rad. These results provide strong evidence of the success of our balance controller design.
It's worth noting that the slight deviation of $\theta_d$ from 0 in this context is attributed to the mounting error of the head IMU. Despite this negligible offset, our balance controller design effectively achieves the desired pitch angle control and demonstrates excellent performance in real-world prototypes.

\begin{figure}[t]
    \centering
    \includegraphics[width = 1.0\columnwidth]{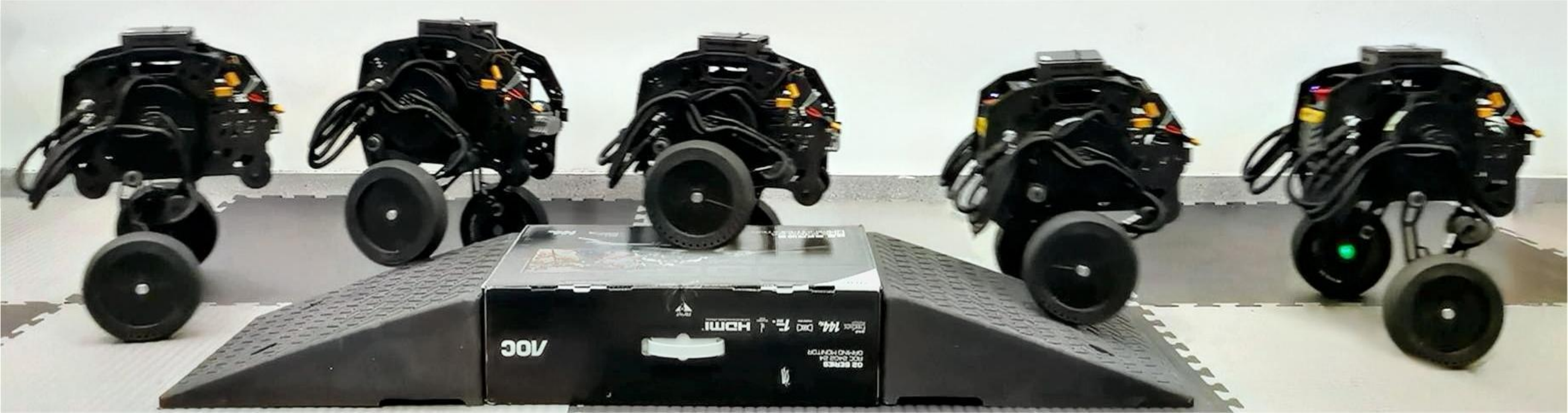}
    \caption{Experiment of roll controller and height controller: single leg obstacle crossing}
    \label{fig:DIABLO's pose performance when passing a slope}
   
    \centering
    \includegraphics[width = 1.0\columnwidth]{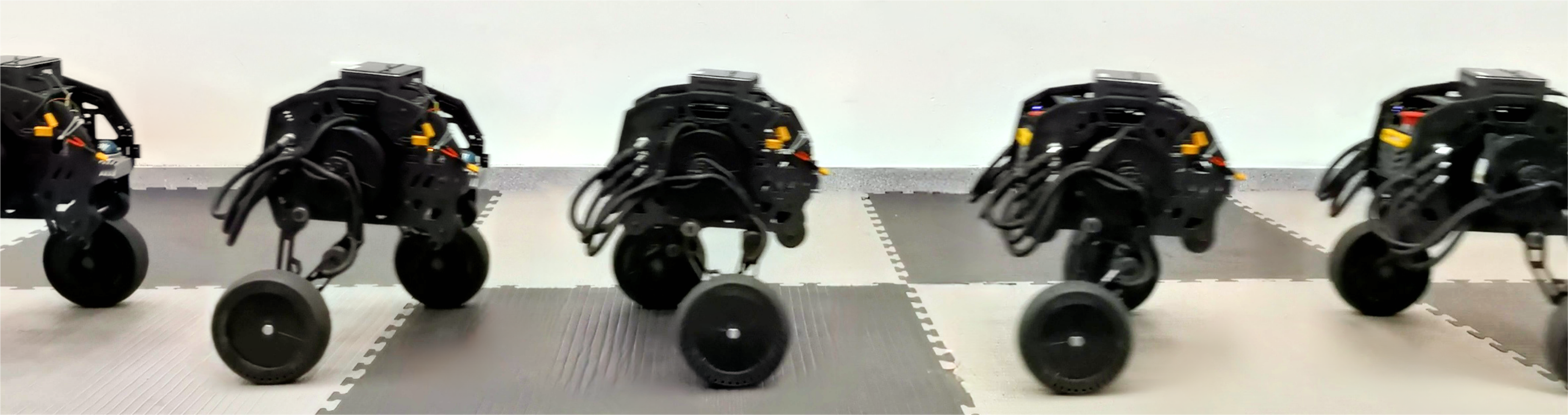}
    \caption{Experiment of split-angle controller: spacewalk}
    \label{fig:DIABLO's pose performance during the spacewalk}
    \vspace{-0.4cm}
\end{figure}

To evaluate the effectiveness of the height controller and roll controller, we constructed a 30° slope (approximately 0.523 rad). DIABLO's pose performance was observed as it traversed the slope on a single leg.
Fig.~\ref{fig:DIABLO's pose performance when passing a slope} showcases its remarkable agility in adapting to changes in slope height while maintaining forward motion and a horizontal orientation for the head. The controller dynamically adjusts leg length for stability during operation, maintaining roll angle variation within 0.1 rad throughout traversal. This successful integration of height and roll controllers enables stable motion on slopes. 

The split-angle controller, initially designed to address model mismatch issues, typically operates with a target input set to zero during regular motion, making it difficult to observe the effectiveness of the controller. However, in specific gaits such as the spacewalk demonstrated in the Fig.~\ref{fig:DIABLO's pose performance during the spacewalk}, the controller's input is a non-zero sinusoidal signal.
During the spacewalk gait, DIABLO achieves a maximum split leg angle of 0.5 rad while maintaining a forward speed of 2 m/s, validating the effectiveness of the split-angle controller for special gaits.
The successful execution of the spacewalk gait, with significant split leg angles while maintaining stability and desired forward speed, highlights the efficacy of our split-angle controller. This demonstrates the controller's adaptability to specific gait patterns and confirms its role in compensating for model mismatch issues.

The conducted experiments serve to validate both the accuracy of the dynamic model presented in Section \ref{sec:modeling} and the reliability of the controllers developed in Section \ref{sec:controller design}.
These results provide DIABLO, the first fully direct-driven wheeled bipedal robot, with robust mobility.

\section{CONCLUSION AND FUTURE WORK}
\label{sec:conclusion}
This paper introduces the DIABLO platform, a WBR composed entirely of direct-drive joints. The utilization of direct-drive motors offers numerous advantages, such as eliminating transmission structure losses and achieving a high control bandwidth. By simplifying the robot to a 2D simplified model, we derived its dynamics equation, enabling a deeper understanding of its motion characteristics.
Furthermore, a model-based LQR controller has been successfully implemented on the prototype. With the assistance of the comprehensive motion controller, DIABLO demonstrates its stability and balance even in scenarios involving body tilt and non-horizontal head orientation(as shown in Fig.~\ref{fig: A novel 6-DoF wheeled bipedal robot}B and Fig.~\ref{fig:DIABLO'S attitude and data during motion in real-wold prototype}A). Through multiple experiments conducted with the prototype system, we illustrate the robot's adaptability and locomotion capabilities across various terrains.

In our future work, we intend further to explore the potential of direct-drive motors for wheel-legged robots. Besides, we plan to implement whole-body control (WBC) methods for controlling different locomotion tasks. These advancements will contribute to enhancing the versatility and performance of DIABLO, pushing the boundaries of wheeled bipedal robotics.

\end{document}